\documentclass[letterpaper, 10 pt, conference]{ieeeconf}  

\IEEEoverridecommandlockouts                              

\usepackage{caption}
\usepackage[ruled]{algorithm2e}
\newcommand{\nextnr}{\stepcounter{AlgoLine}\ShowLn}
\usepackage{algpseudocode}
\algnewcommand\algorithmicforeach{\textbf{for each}}
\algdef{S}[FOR]{ForEach}[1]{\algorithmicforeach\ #1\ \algorithmicdo}
\SetKwRepeat{Do}{do}{while}

\usepackage{booktabs} 
\usepackage{graphicx} 
\usepackage{makecell} 
\usepackage{amsmath} 
\usepackage{amssymb} 
\usepackage{xcolor,blindtext} 
\usepackage{subfig}
\usepackage{color} 
\usepackage{balance} 
\usepackage{multirow}

\begin{document}

\title{SLAM assisted 3D tracking system for laparoscopic surgery}

\author{Jingwei Song, Ray Zhang, Wenwei Zhang, Hao Zhou, and Maani Ghaffari%
\thanks{J. Song is with United Imaging Research Institute of Intelligent Imaging, Beijing 100144, China
 \texttt{jingweisong.eng@outlook.com}}%
 \thanks{R. Zhang and M. Ghaffari are with the University of Michigan, Ann Arbor, MI 48109, USA. \texttt{\texttt{\{rzh, maanigj\}@umich.edu}.}}
 \thanks{H. Zhou is with United Imaging, Shanghai, China. \texttt{Hao.Zhou@united-imaging.com}. }
  \thanks{W. Zhang is with United Imaging of Surgery, Shanghai, China. \texttt{wenwei.zhang@uih-surgical.com}. }
}
\maketitle
\thispagestyle{empty}
\pagestyle{empty}

\begin{abstract}

A major limitation of minimally invasive surgery is the difficulty in accurately locating the internal anatomical structures of the target organ due to the lack of tactile feedback and transparency. Augmented reality (AR) offers a promising solution to overcome this challenge. Numerous studies have shown that combining learning-based and geometric methods can achieve accurate preoperative and intraoperative data registration. This work proposes a real-time monocular 3D tracking algorithm for post-registration tasks. The ORB-SLAM2 framework is adopted and modified for prior-based 3D tracking. The primitive 3D shape is used for fast initialization of the monocular SLAM. A pseudo-segmentation strategy is employed to separate the target organ from the background for tracking purposes, and the geometric prior of the 3D shape is incorporated as an additional constraint in the pose graph. Experiments from in-vivo and ex-vivo tests demonstrate that the proposed 3D tracking system provides robust 3D tracking and effectively handles typical challenges such as fast motion, out-of-field-of-view scenarios, partial visibility, and ``organ-background'' relative motion.

\end{abstract}


\IEEEpeerreviewmaketitle

\section{Introduction}

\begin{figure}[t]
    \centering
    \includegraphics[width=0.9\columnwidth]{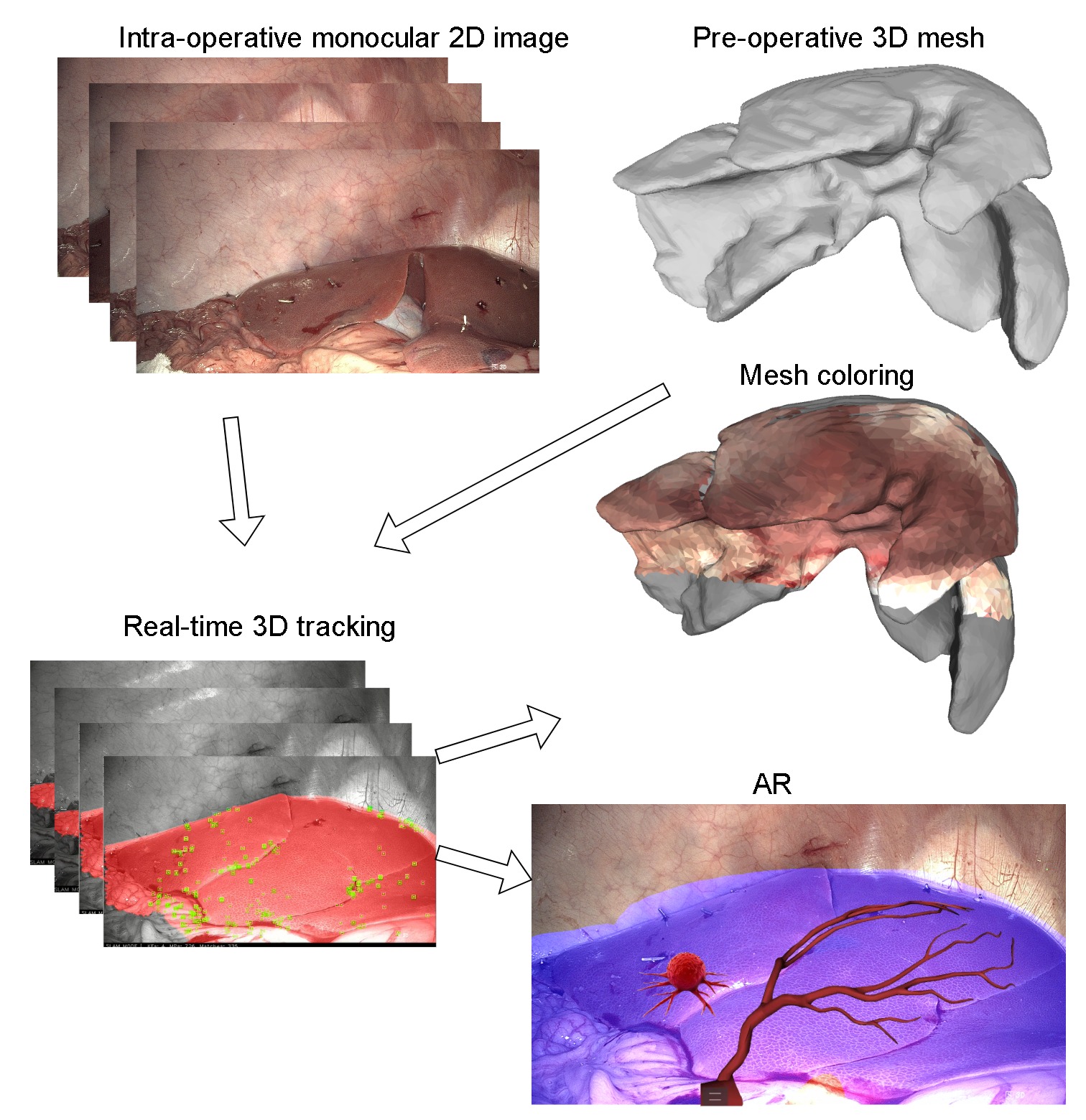}
    \caption{The figure shows the pre-operative 3D images, intra-operative registration procedure, and potential clinical applications.}
    \label{fig_AR}
\end{figure}

Robot-assisted laparoscopic surgery involves inflating the abdominal cavity with gas and inserting a rigid laparoscope through a trocar in the abdominal wall to visualize the peritoneal cavity contents~\cite{monnet2003laparoscopy}. Robotic guidance allows surgeons to manipulate instruments with minimal tremors, reduced fatigue, and shorter training times~\cite{alkatout2021development,pore2023autonomous}. However, surgeons face challenges in localizing internal structures, such as tumors and vessels, due to the inability to use hand palpation. To address this, Augmented Reality (AR)-assisted approaches~\cite{bourdel2017use,collins2020augmented} incorporate pre-operative imaging data, like Computed Tomography (CT) or Magnetic Resonance Imaging (MRI) scans, to help localize internal structures during robot-assisted Minimally Invasive Surgery (MIS). AR facilitates simultaneous surface and subsurface visualization in laparoscopic videos, enabling surgeons to correlate both in real-time imagery~\cite{pessaux2014robotic,pessaux2015towards}. With accurate, real-time alignment of pre-operative 3D images and intra-operative video, the internal anatomy can be visualized by overlaying it onto the surgical video feed. Fig. \ref{fig_AR} shows the general procedure of the 3D to 2D registration and its potential clinical applications.\par 

A key component of AR-guided laparoscopic navigation is achieving accurate 3D-to-2D registration, where 3D refers to pre-operative 3D imaging (segmented as mesh) and live laparoscopic 2D  video. Some approaches use external tracking devices with optical markers to perform registration~\cite{luo2020augmented,pelanis2021evaluation}. While these systems offer robust registration, their application is limited due to the invasiveness of marker placement, the complexity and cost of deployment, marker occlusion, and challenges with tracking tissue deformation~\cite{zhang2019markerless,malhotra2023augmented}. Consequently, markerless registration has gained popularity in recent research.\par

Markerless registration algorithms typically follow a two-stage process: coarse (global) and fine (local) registration~\cite{modrzejewski2019vivo,guo2020deep}, similar to point cloud registration in computer vision society~\cite{deng2022survey}. Coarse registration aligns the pre-operative 3D model with intra-operative 2D images at the start of surgery when deformation and topology remain relatively constant. During this phase, surgeons capture a large field of view (FoV) of the target organ. Based on this alignment, real-time tracking ensures low-latency registration for AR or robotic applications. Fine registration, also known as 3D tracking, iteratively refines the alignment between 3D and 2D video in real time. Fine registration, or 3D tracking, leverages the initial coarse registration and uses lightweight models for fast and accurate registration.\par



Coarse (global) registration has been extensively studied within the research community. It is widely accepted that fusing semantic information~\cite{hayoz2023learning,karaoglu2023ride} with anatomical contour-based constraints~\cite{koo2022automatic} yields robust 3D-to-2D multi-modal registrations. However, Deep Neural Network (DNN)-based methods, while effective, require substantial computational resources and are unable to ensure real-time 3D tracking. A pioneering tracking algorithm developed by~\cite{bourdel2017use} enabled real-time visualization of myomas in gynecological surgery. Inspired by this work, we demonstrate that Simultaneous Localization And Mapping (SLAM) can significantly improve 3D tracking by providing more robust and accurate pose estimation. In particular, the pose graph module enhances tracking accuracy and robustness against occlusion and organ movement. Additionally, loop closure allows for re-localization after tracking loss. Our system offers a cost-effective solution for modern AR-based laparoscopy navigation.

The contributions of this article are as follows:
\begin{itemize}
    \item We propose a monocular, real-time 3D-to-2D tracking system for laparoscopic surgery.
    \item The learning-based 3D-to-2D coarse registration algorithm is combined with geometric constraints for improved accuracy.
    \item Primitive 3D shape is leveraged to provide a pseudo segmentation between the target organ and background, and it also serves as a pose graph constraint to enhance SLAM performance.
    \item We propose a new texturing strategy for primitive 3D shape visualization.
\end{itemize}

\section{Literature review}

Based on the existence of an initial registration, approaches can be categorized into coarse (global) registration and fine (local) registration, also known as 3D tracking~\cite{guo2020deep}.

Coarse registration focuses on determining the rigid spatial transformation between the pre-operative 3D model and the intra-operative 2D images, which exist in separate coordinate systems. As it is typically used for initialization or re-initialization in AR, speed is less critical than robustness. Approaches can be semi-automatic or fully automatic~\cite{espinel2020combining}. The primary challenge in this phase is that 3D geometry and 2D textured images are in different domains. Semi-automatic approaches involve user interaction for manual associations, using visual cues, semantic regions, and ridges~\cite{espinel2020combining,nicolau2011augmented,collins2020augmented}, after which 3D-2D registration algorithms estimate the optimal 3D pose. Automatic methods handle data associations without user input. For instance, \cite{koo2022automatic} used inferior ridges for contour-based registration, while \cite{espinel2021using} incorporated contours, silhouettes, inferior ridges, and semantic information, such as the Falciform ligament, for segmentation. Unlike methods that rely on geometric modeling, \cite{labrunie2023automatic} proposed an end-to-end iterative search for optimal rigid and non-rigid transformations. Similarly, \cite{karaoglu2023ride} used a rotation-equivariant neural network for invariant descriptor extraction, and \cite{hayoz2023learning} combined semantic and geometric information for 3D-2D stereo video matching. Utilizing 3D partial scans recovered from stereoscopic imagery, \cite{robu2018global} applied the TOLDI geometric descriptor~\cite{yang2017toldi} for 3D model alignment with stereo videos.\par 

Fine registration, or 3D-2D tracking, is the process of continuously aligning the pre-operative 3D model with the intra-operative 2D images following coarse registration. Fine registration requires fast computations for clinical applications. Prior work~\cite{bourdel2017use,bourdel2019use,collins2020augmented,song2021combining} relied on visual cues from sequential 2D video frames to recover the 3D trajectory of a moving laparoscope. The 6 Degrees of Freedom (DoF) poses are estimated using tracked geometric features such as points, curves, and landmarks, which are considered either rigid or non-rigid. Once one or more images in the video are aligned with the pre-operative 3D data during coarse registration, the entire recovered trajectory can be aligned with the pre-operative data. Different approaches vary in the types of features used and their assumptions about rigidity. For example, \cite{bourdel2017use,bourdel2019use,collins2020augmented,song2021combining} employed corner points and keyframes for rigid tracking, while \cite{collins2020augmented} combined corner points and silhouettes for both rigid and non-rigid tracking. \cite{labrunie2022automatic} segmented the inferior ridge and Falciform ligament using DNNs for both rigid and non-rigid registration, while \cite{fu2023novel} used contours for 2D-3D registration in intrarenal multimodal navigation. \cite{wang2024video} applied a probabilistic rigid registration and elastic compensation algorithm for 3D-2D registration, using a stereoscope instead of a monocular system. To date, no non-rigid registration algorithm has demonstrated real-time performance.


  \begin{figure*}[!h]
		\centering
		\subfloat{
			\begin{minipage}[]{0.72\textwidth}
				\centering
				\includegraphics[width=1\linewidth]{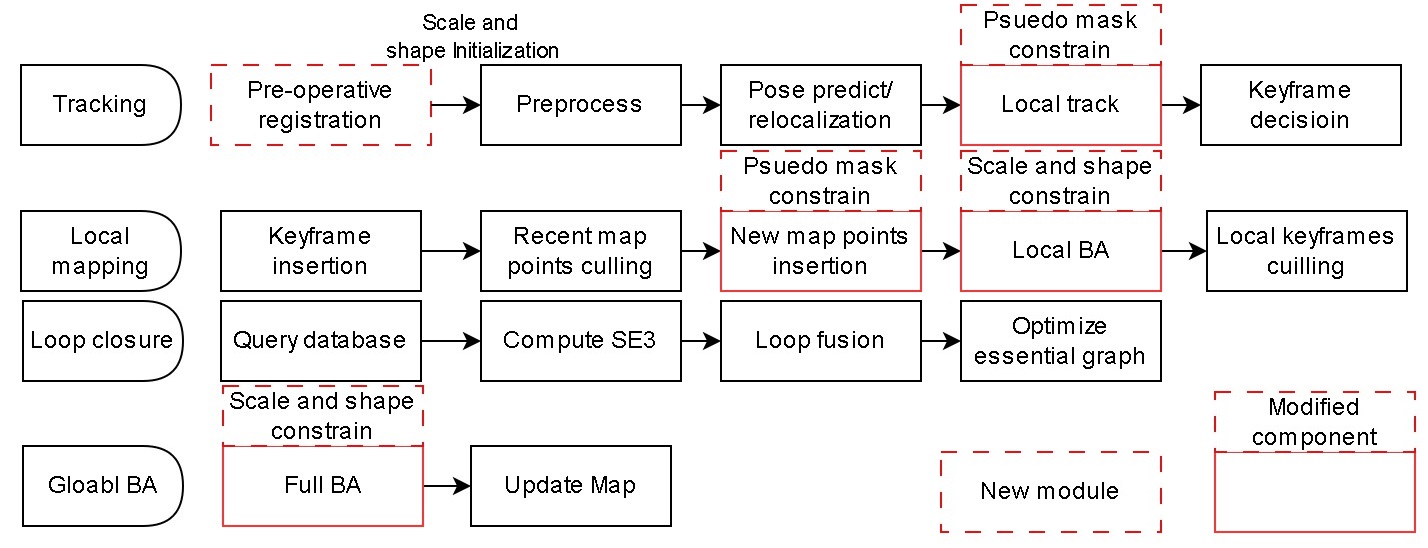}
			\end{minipage}
		}
		\caption{This figure presents the overall pipeline of the proposed method and also serves as a comparison between our approach and ORB-SLAM2.}
		\label{fig_pipeline}
	\end{figure*}

	\section{Methodology}

\subsection{System overview}

The problem is formulated as a special monocular SLAM task, where a prior 3D mesh is correctly registered to the first frame but lacks texturing. Fig. \ref{fig_pipeline} illustrates the pipeline of our proposed method. Our 3D tracking algorithm is built on ORB-SLAM2~\cite{mur2017orb}, with additional modules allowing integrating prior 3D shape for robust 3D tracking. Modifications and adaptations are highlighted in red in Fig. \ref{fig_pipeline} and will be discussed in detail in the following sections. Broadly, the tracking system follows the parallel ``tracking'' and ``mapping'' paradigm, incorporating global Bundle Adjustment (BA) and loop closure to minimize drift and recover from tracking loss. The key adaptations in this work focus on leveraging the prior 3D shape to reduce scale drift, account for relative motion between the target shape, and enhance mapping and localization accuracy.

3D tracking in Laparoscopic surgery differs from monocular SLAM in the following perspectives 
\begin{itemize}
    \item Scale should be estimated in a 3D tracking system, while monocular SLAM cannot retrieve scale.
    \item There is a relative motion between the target organ and the background. \textbf{The coordinate system should be built on the pre-operative shape.}
    \item The prior map (3D prior shape) is known. Yet, it only preserves geometry, which is in a different modal from the image.    
\end{itemize}

The following sections handle these issues. While this work considers fast motion, objects moving out of the FoV, partial visibility, and ``organ-background'' relative motion, other issues, including deformation, fluorescence, bad image quality, blood, and fog, are ignored.

\subsection{Registration and initialization with prior shape}

Define $\mathbf{T}_{init} \in SE3$ as the global registration, which aligns the 3D pre-operative mesh to the first frame of the intra-operative image. This research obtains the $\mathbf{T}_{init}$ from the semi-automatic registration. The transformation is used to initialize the monocular SLAM system. The coordinate system of our AR system is in a pre-operative 3D image. 

\begin{equation}
\label{eq_sfm}
\mathbf{T}_{init} = 
\operatorname{argmin}_{\mathbf{T}} \sum_{i \in \Gamma}||\pi(\mathbf{T}\mathbf{f}_i,\mathbf{K})-\mathbf{q}_{i}||_2,
\end{equation}

\noindent where $\mathbf{f}_i \in \mathbb{R}^4$ is the 3D point in homogeneous coordinate and $\mathbf{q}_{i} \in \mathbb{R}^3$ is the corresponding 2D point on the first image. $\Gamma$ is the set of all selected points indexes. $\pi(\mathbf{T}\mathbf{f}_i,\mathbf{K}) = \mathbf{K}\mathbf{T}\mathbf{f}_i$ projects 3D point on 2D plane based on the scope's intrinsic matrix $\mathbb{R}^{3\times4}$. A branch-and-bound strategy~\cite{olsson2008branch} is applied to the pose initialization for better convergence. The solution with the minimal sum of residuals is selected. It is worth noting that the transformation can also be automatically obtained using semantic information from a DNN-based segmentation network~\cite{bourdel2017use,collins2020augmented}.

The transformation $\mathbf{T}_{init}$ and the 3D prior shape are used to initialize the SLAM system. The 3D prior shape is obtained from the pre-operative 3D images, such as CT or MRI, which are segmented and converted into the mesh format. Unlike traditional monocular SLAM, which initializes by performing BA on the first few images, our approach projects the 3D prior shape onto 2D and initializes the system based on the ``simulated depth''. This offers two key advantages. First, BA-based initialization requires significant translational parallax between images~\cite{liu2020parallax}, which can be burdensome for surgeons. In contrast, prior shape-based initialization ensures robust performance with just the first frame. Second, the ambiguous scale can be accurately recovered.

\begin{figure}[t]
    \centering
\includegraphics[width=0.85\columnwidth]{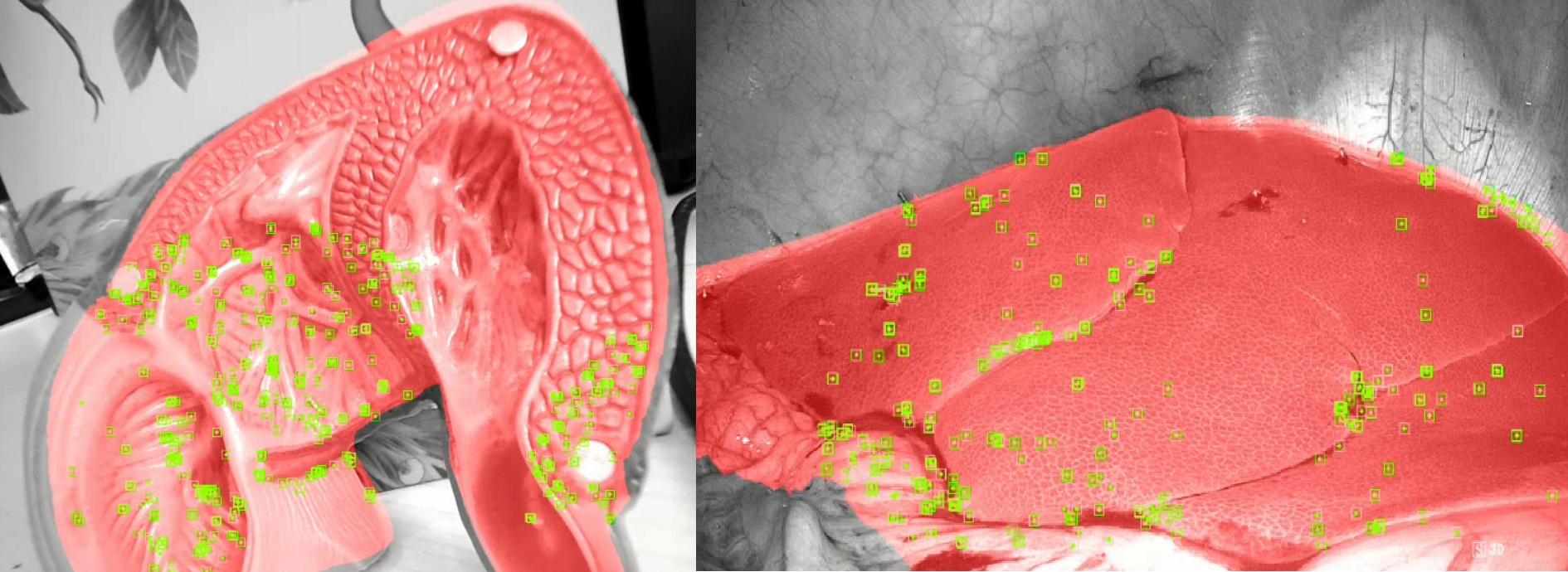}
    \caption{The figure illustrates an example of pseudo segmentation and key point tracking. The red mask represents the pseudo-segmented target organ, while the green points are the tracked key points within the mask. Corner points outside the mask are filtered out.}
    \label{fig_sample_seg}
\end{figure}
\subsection{Pseudo target segmentation and tracking}

A major challenge in laparoscopy is the significant relative motion of the target organ against the background during surgery. The operators deliberately manipulate the target organs, making the ``organ-background'' relative motion a larger factor than organ deformation in terms of tracking performance (organ manipulations can be found at~\cite{aragon2012techniques} and open-source videos). By focusing tracking and mapping \textbf{only on the target organ}, robustness can be significantly enhanced. A simple approach is to use a lightweight DNN to semantically segment the target organ from the image. Inspired by the fast semi-automatic process with a touchscreen interface from~\cite{collins2020augmented}, we propose a method of pseudo-masking that is prior-free.

Pseudo-segmentation for time $i$ is generated by projecting the prior shape in the pose of frame $i-1$. The key assumption is that the trajectory of sequential frames is smooth, and the tracking of the last frame is relatively accurate. Once the mask is created, SLAM is performed within the mask, meaning that all key points are selected from within the masked region. This allows for a better estimation of the spatial relationship between the monocular camera and the target organ. 

Fig. \ref{fig_sample_seg} illustrates the pseudo mask-based filtering procedure. The red mask represents the projected 3D shape prior from the previous frame. This mask is used to filter out corner points that fall outside its boundary. The green points, which lie within the mask, are obtained and used for 3D tracking of the target organ.

\subsection{Pose graph with prior geometry}

 As the prior shape is available, its geometry can also be used as a constraint for rectifying pose drift. This work incorporates prior shape in both ``Global BA'' and ``Local BA'' modules in ORB-SLAM2. ``Global BA'' optimizes the all keyframe poses and map points while ``Local BA'' optimizes local keyframe pose (ones pose) and its associated map points. The prior shape is used as a soft constraint for map points in ORB-SLAM2. The authentic ``Global BA'' formulation~\cite{kummerle2011g,carlone2018convex} is
\begin{equation}
\label{eq_global_BA}
\begin{aligned}
&\mathbf{T}_2^*,...,\mathbf{T}_n^*,\mathbf{f}^*_1,...,\mathbf{f}^*_m = \\
&\operatorname{argmin}_{\mathbf{T}_2,...,\mathbf{T}_n,\mathbf{f}_1,...,\mathbf{f}_m} \sum_{i \in \Omega,j \in \Gamma}||\pi(\mathbf{T}_i\mathbf{f}_j,\mathbf{K})-\mathbf{q}_{ij}||_2,
\end{aligned}
\end{equation}

\noindent where $\mathbf{T}_2^*,...,\mathbf{T}_n^*$ are all poses of key frames ($\mathbf{T}_1^*$ is fixed as $\mathbf{T}_{init}$), $\{\mathbf{f}^*_j\}$ are all sparse 3D map points and $\{\mathbf{q}^*_{ij}\}$ is their 2D projection on image $i$. $\Omega$ and $\Gamma$ are the index set for poses and points. 

\begin{figure}[t]
    \centering
    \includegraphics[width=0.98\columnwidth]{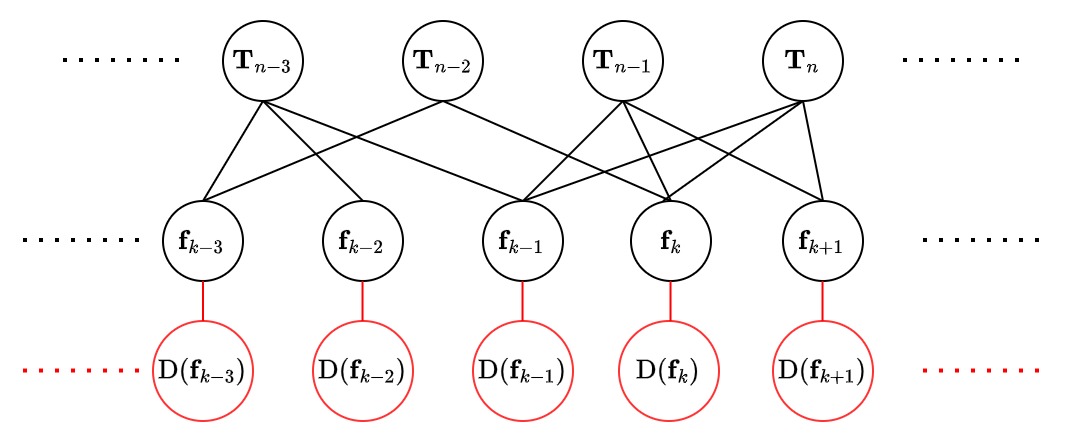}
    \caption{The figure shows the pose graph which is constrained with the prior shape. The red circles are the closest point on the prior shape.}
    \label{fig_pose_graph}
\end{figure}

\eqref{eq_global_BA} is formulated and solved based on pose graph. Vertices include all poses and 3D feature points. 3D to 2D projection serves as the edges connecting poses and feature points. Applying the prior shape as constraint, closest point on the shape is calculated regarding the feature point and added to the pose graph. Their distance is set as the edge. Fig. \ref{fig_pose_graph} shows the corresponding pose graph. \eqref{eq_global_BA} is reformulated as 

\begin{equation}
\label{eq_sfm_priorshape}
\begin{aligned}
&\mathbf{T}_2^*,...,\mathbf{T}_n^*,\mathbf{f}^*_1,...,\mathbf{f}^*_m =\\
&\operatorname{argmin}_{\mathbf{T}_2,...,\mathbf{T}_n,\mathbf{f}_1,...,\mathbf{f}_m} \sum_{j \in \Gamma'}
\rho(\Vert\pi(\mathbf{T}_i\mathbf{f}_j,\mathbf{K})-\mathbf{q}_{ij}\Vert_2) +\\ &\mathrm{w}_{shape}\rho(\Vert\operatorname{D(\mathbf{f}_i)}-\mathbf{f}_i\Vert_2),
\end{aligned}
\end{equation}

\noindent where $D(\mathbf{f}_i)$ locates the closest point of $\mathbf{f}_i$ on the prior shape. $\rho(\cdot)$ is the M-estimator. $\mathrm{w}_{shape}$ is hyper-parameter. To fasten the closest point searching, the prior shape is converted into dense point clouds and K-D tree~\cite{muja2009flann} is adopted for fast searching.  The same constraint also applies to ``local BA'' module. 



\subsection{Prior shape texturing}

This work follows \cite{song2018mis} by coloring the prior textureless shape with weighting. Denote $\Delta_i$ as the triplet of the 3 vertices of the triangle. The color of $\Delta_i$ is $\mathbf{c}_i \in R^3$. The weight for $\Delta_i$ is denoted $\mathrm{w}_i$. Triangle's normal $\mathbf{n}_i \in R^4$ is in homogeneous coordinate. Algorithm \ref{Algorithm_shape_texture} presents the procedure of the texture coloring.

\begin{algorithm}[t]
	\caption{Prior shape texturing.}
	\label{Algorithm_shape_texture}
	\KwIn{Triangle triplet $\Delta_i$, color $\mathbf{c}_i$, normal $\mathbf{n}_i$ $\mathrm{w}_i$, $\mathbf{T} \in SE3$ and image $\mathbf{I}$} 
	\KwOut{$\mathbf{c}_i$}
    \nextnr
	$\mathrm{i} = 1$\\
    \nextnr
    \While{$\mathrm{i} < \mathrm{I}$}{
        \nextnr
        \If{$(\mathbf{T}\mathbf{n})|_z<0$}
        {
            $\mathbf{c}'=\Pi(\Delta_i,\mathbf{T},\mathbf{I},\mathbf{K})$\\
            \nextnr
            $\mathbf{c}_i = \frac{\mathbf{w}_i\mathbf{c}_i+\mathbf{c}'}{\mathbf{w}_i+1}$\\
            \nextnr
            $\mathbf{w}_i = \operatorname{max}(10,\mathbf{w}_i+1)$\\
        }
	}
	Notation: $\mathrm{I}$ is the number of triangles. $\Pi(\Delta_i,\mathbf{T},\mathbf{I},\mathbf{K})$ obtains the average color on image $\mathbf{I}$ of the projected triangle $\Delta_i$ based on the selected pose $\mathbf{T}$ and intrinsic parameter $\mathbf{K}$. $\mathbf{n}_i|z$ extract the $z$ value in the normal.\\
\end{algorithm}

\begin{figure}[t]
    \centering
\includegraphics[width=1\columnwidth]{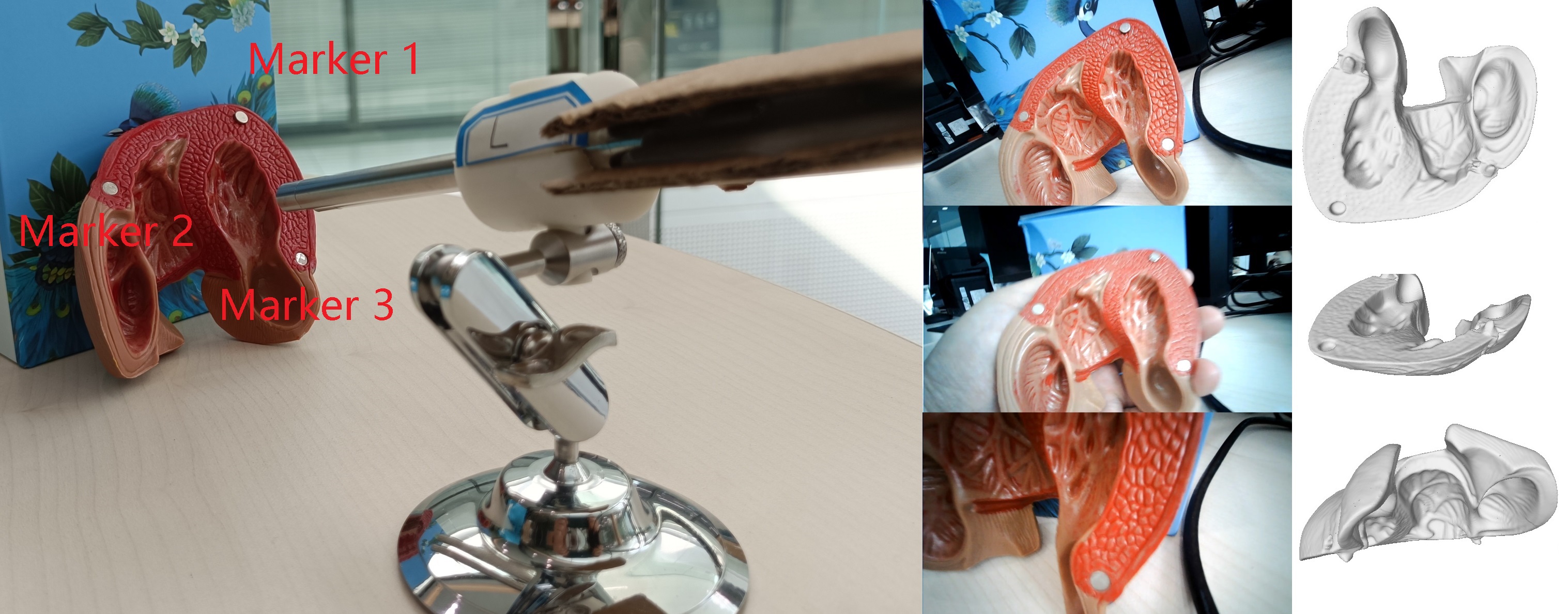}
    \caption{The figure shows our setup of the ex-vivo data set collection. From left to right are the hardware setup, sample images, and phantom's 3D mesh collected by ourselves. }
    \label{fig_ex_vivo}
\end{figure}

\begin{figure*}[!h]
		\centering
		\subfloat{
			\begin{minipage}[]{0.92\textwidth}
				\centering
				\includegraphics[width=1\linewidth]{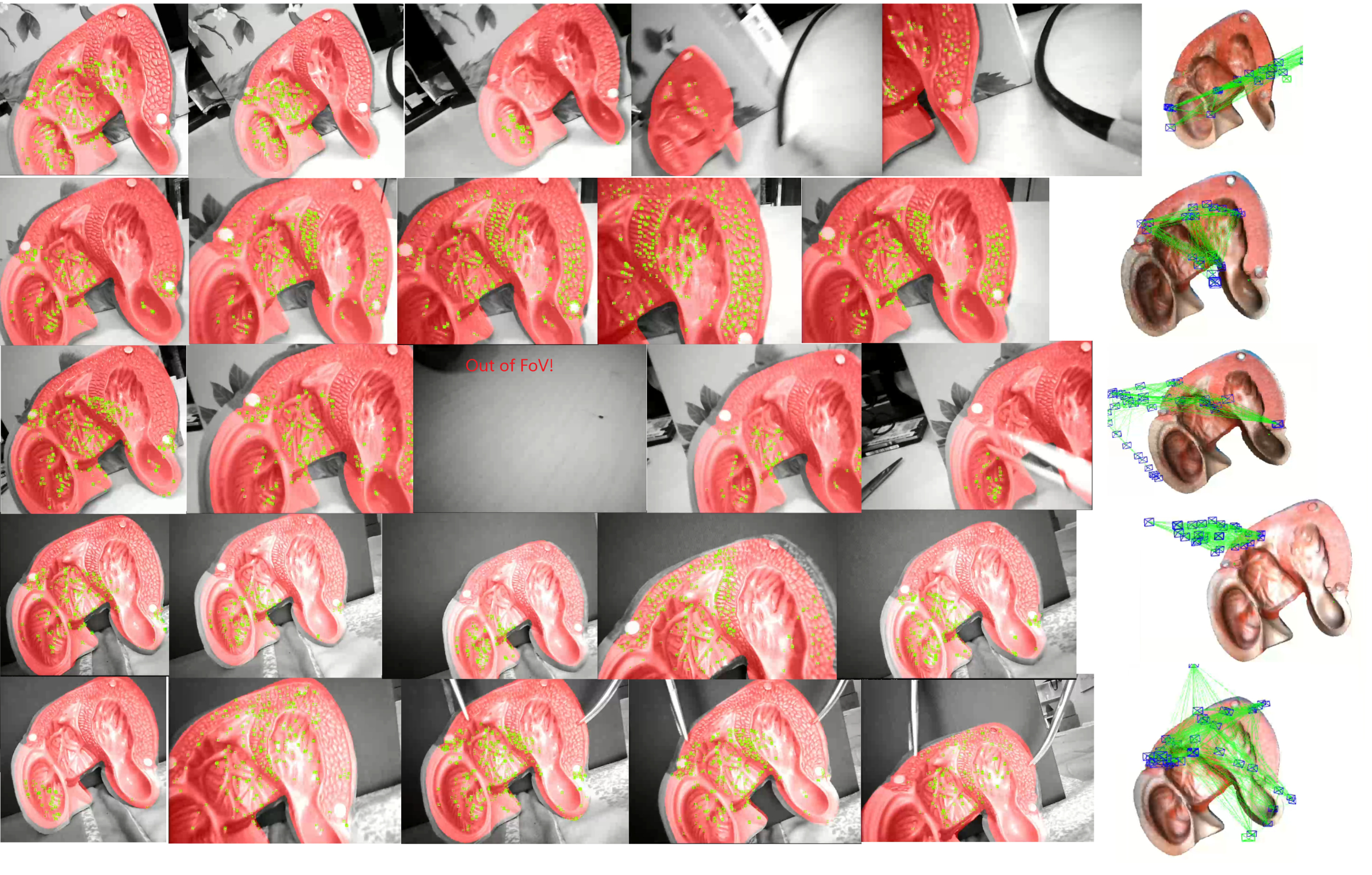}
			\end{minipage}
		}
		\caption{The figure shows sample 3D tracking results of the proposed method on the phantom. Each row represents one sequences. The last column is the 3D mesh colored by 3D tracking system.}
		\label{fig_results_phantom}
	\end{figure*}

 \begin{figure*}[!h]
		\centering
		\subfloat{
			\begin{minipage}[]{0.92\textwidth}
				\centering
				\includegraphics[width=1\linewidth]{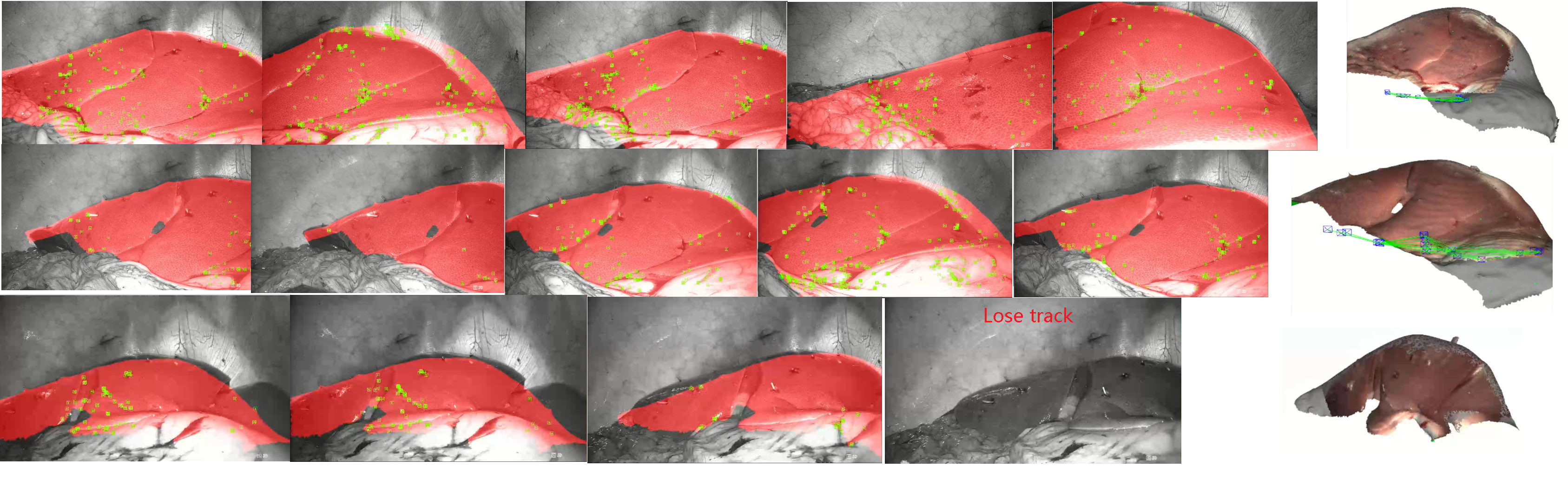}
			\end{minipage}
		}
		\caption{The figure shows sample 3D tracking results of the proposed method on the DePoLL data set. Each row represents one sequences. The last column is the 3D mesh colored by 3D tracking system.}
		\label{fig_results_invivo}
	\end{figure*}

	\section{Results and discussion}

    The proposed AR system was implemented based on ORB-SLAM2~\cite{mur2017orb}, with various adaptations and modifications (all in C++). It was deployed and tested on a desktop equipped with an i7-12700 CPU, 32 GB of memory, and an RTX 3060 GPU. The system runs primarily on the CPU, using approximately 3-5 GB of memory during all tests. The hyperparameter $\mathrm{w}_{shape}$ was set to $100$, while other hyperparameters followed the default settings from~\cite{mur2017orb}.

    The in-vivo dataset, DePoLL (Deformable Porcine Laparoscopic Liver)~\cite{modrzejewski2019vivo}, collected by the lab IRCAD, contains 13 intraoperative stereo videos, metal markers, and reference 3D models segmented from intraoperative CT scans. The videos are about 1 minute long with a frame rate of 50 Hz. We decoded and resized them to $960 \times 540$. Only the left video was used for evaluation. A major limitation of the dataset is the absence of an intrinsic camera matrix, so we used it only for qualitative evaluation. During our experiments, we found that setting the focal length to 700 and the optical center to $[480, 270]$ was sufficient to achieve the initial registration.

    An ex-vivo dataset was collected using a monocular scope to capture sequential images of the phantom. Fig. \ref{fig_ex_vivo} shows the experimental setup and sample images. The 3D CT images were obtained using United Imaging's uCT 860, and the United Imaging's commercial software was used to segment the 3D mesh. During image collection, typical challenges such as fast motion, objects moving out of the field of view, partial visibility, and ``organ-background'' relative motion were introduced and tested. The image resolution was $1280 \times 720$, and all images were calibrated and initially aligned with the pre-operative mesh by computing $\mathbf{T}_{init}$. A total of five image sequences were collected, each containing over 4,000 images.\par

    The proposed algorithm was evaluated on time consumption, qualitative performance, accuracy, and an ablation study. DePoLL was used for qualitative test only since no intrinsic parameter was provided. Meanwhile, our ex-vivo data set has been well-calibrated. The plastic phantom has three metal circle markers (shown in Fig. \ref{fig_ex_vivo}), and we use them for manual evaluation. Specifically, the 2D Target Registration Error (TRE) is measured for performance evaluation.

\subsection{3D tracking performance}

Fig. \ref{fig_results_phantom} and Fig. \ref{fig_results_invivo} show sample 3D tracking results of the proposed method on the ex-vivo phantom and in-vivo porcine (DePoLL dataset). The colored 3D phantom meshes are also visualized\footnote{We highly recommend watching the attached video for a more comprehensive understanding of the performance.}. The left column displays the initial registration. Both the results and the video show the system's ability to handle challenges such as fast motion, out-of-field-of-view (FoV) scenarios, partial observations, and \textit{organ-background} relative motion. 1)  Blurry images caused by fast motion are not trackable, but this has a minimal impact as the endoscope rarely experiences sustained rapid movement. 2) Although the scope occasionally moves out of the FoV during surgery, the loop-closure model in ORB-SLAM2 successfully recovers the pose. 3) The proposed method remains robust even in cases where the scope partially observes the target organ or is occluded by surgical tools, as long as corner point tracking is reliable. 4) The method effectively handles ``organ-background'' relative motion, which occurs when the surgeon manipulates the phantom with forceps. Since the coordinate system is built on the preoperative shape and the pseudo mask filters out the background, the background motion does not affect the 3D tracking performance.

The closest related research is \cite{bourdel2017use} and its follow-up \cite{bourdel2019use}, which also achieved real-time 3D tracking. However, the authors only stated ``This stage runs in real-time and aligns the preoperative models to the live laparoscopic video'' without detailing their tracking method making it impossible for re-implementation. In our ex-vivo dataset, we randomly selected 20 frames with tracking maintained and measured the TRE based on the three circle markers. The average TRE was 5.39 pixels. Table~\ref{Table_exvivo_dataset} summarizes all the quantitative results on the ex-vivo data set.

In Fig. \ref{fig_results_invivo}, drifts increase as the scope moves away from the initial FoV and decreases when the initial FoV is restored. This drift is primarily due to the incorrect intrinsic matrix in DePoLL. In other words, the initial registration only aligns with our ``guessed'' intrinsic matrix in the initial frame. This issue did not arise in our ex-vivo dataset. Overall, the results demonstrate that our proposed system performs well in local tracking.

The system achieves a tracking rate of 26 Hz and 31 Hz in the ex-vivo and in-vivo experiments, respectively, with prior shapes consisting of approximately $20000$ faces. All trajectories are successfully initialized in the first frame. Compared to ORB-SLAM2 backbone, projecting the 3D prior shape onto the 2D image plane adds some computational time. In this preliminary work, each face is sequentially projected on CPU-end. This process can be significantly accelerated using OpenGL on the GPU or simplifying faces. Other modifications impose a minimal computational burden.

\subsection{Ablation study}
\subsubsection{Initialization with prior shape}

The proposed initialization strategy achieves a $100\%$ success rate across all tested data sets. Our shape-based approach functions similarly to RGB-D initialization. In contrast, the standard monocular initialization fails in all DePoLL data sets, likely due to an incorrect intrinsic matrix during BA, as well as its dependence on translational motion for initialization. An example can be seen in Fig. \ref{fig_results_init}.

\subsubsection{Registration with prior shape}

We observe that $27.13\%$ frames were in lost track when the prior shape-based registration strategy was disabled comparing to $7.01\%$ frames. This occurred because fewer feature points were selected on the target organ. Additionally, disabling the prior shape-based strategy caused tracking to fail during organ manipulation, as feature points in the background became outliers for 3D tracking.

\subsubsection{Pose graph with prior geometry}

\begin{figure}[t]
    \centering
\includegraphics[width=0.93\columnwidth]{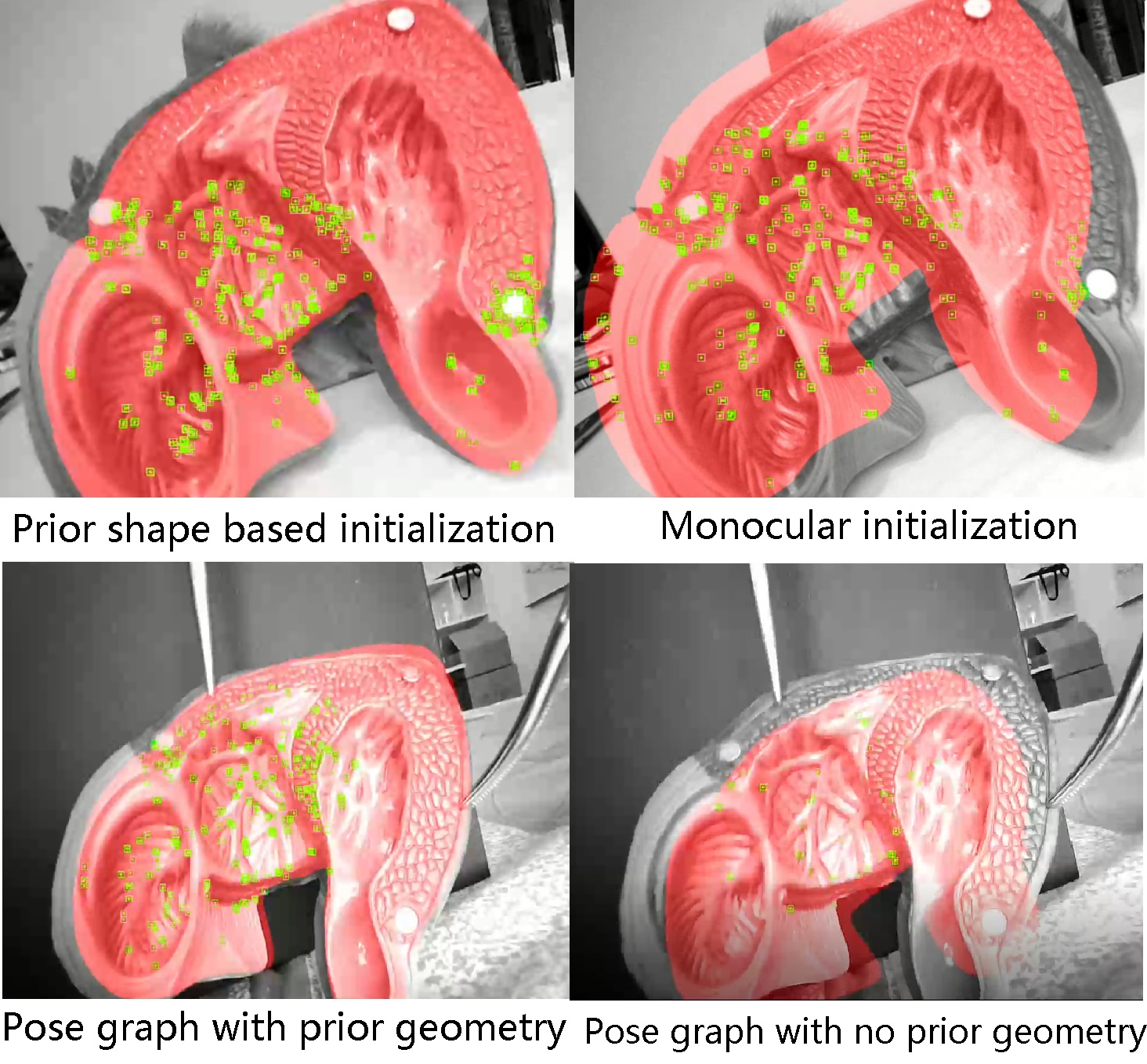}
    \caption{The upper figure shows a comparison between prior shape-based initialization and monocular initialization. The lower figure illustrates a comparison of the pose graph with and without the inclusion of prior geometric information.}
    \label{fig_results_init}
\end{figure}

Results show that pose graph does not provide a noticeable improvement over all 3D trackings. We find few improvements and one is presented in lower figure in Fig. \ref{fig_results_init}. 

\subsection{Limitations}

Although the experiment shows that the proposed AR system is suitable for general 3D tracking and can handle fast motion, out of FoV, partial observation, and ``organ-background'', it can be improved by addressing deformation, fluorescence, bad image quality, blood, and fog. Moreover, it is unfortunate that the DePoLL data set lacks an intrinsic matrix and cannot be applied to perform quantitative assessments. A similar data set will be beneficial for the quantitative evaluation of these 3D tracking systems.

\begin{table}[]
		\centering
		\caption{The table is the quantitative results of the 3D tracking system.}
  
		\begin{tabular}{ccc}
\hline
TRE         & Frame rate & Averge lost frames \\ \hline
5.39 pixels & 26 Hz      & 7.01\%             \\ \hline
\end{tabular}
		\label{Table_exvivo_dataset}
	\end{table}

\section{Conclusion}

This work proposes a real-time 3D tracking method for an AR system in monocular laparoscopic surgery and MIS. The ORB-SLAM2 system is adapted and modified for prior-based multimodal 3D tracking. Specifically, three key modifications have been made. 1. The 3D shape prior is used for fast initialization of the monocular SLAM. 2. A pseudo-segmentation strategy is introduced to segment the target organ from the background, limiting tracking to the segmented organ. 3. the geometric prior of the 3D shape is applied as an additional constraint in the pose graph. Experimental results from both in-vivo and ex-vivo tests demonstrate that the AR system provides robust 3D tracking. The proposed 3D tracking system is being commercialized by the company United Imaging of Surgery.\par

\balance
{\small
		\bibliographystyle{ieeetr}
		\bibliography{bib/strings-abrv,bib/ieee-abrv,annot}
	}

\end{document}